\documentclass[letterpaper]{article} 
\usepackage{aaai2027}
\usepackage[hyphens]{url}  
\usepackage{graphicx} 
\usepackage{natbib}  
\usepackage{caption} 
\usepackage{algorithm}
\usepackage{algpseudocode}

\usepackage{newfloat}
\usepackage{listings}

\DeclareCaptionStyle{ruled}{labelfont=normalfont,labelsep=colon,strut=off} 
\floatstyle{ruled}
\newfloat{listing}{tb}{lst}{}
\floatname{listing}{Listing}

\usepackage{booktabs}

\usepackage{multirow}

\newif\ifpromptwide
\promptwidefalse          

\usepackage[most]{tcolorbox}

\ifpromptwide
  \usepackage{cuted}
\else
  \newenvironment{strip}{}{}
\fi

\definecolor{promptorange}{HTML}{98CADD}
\definecolor{promptbg}{HTML}{EAEFF6}

\ifpromptwide
  \tcbset{promptbreak/.style={}}
\else
  \tcbset{promptbreak/.style={breakable}}
\fi

\newtcblisting{promptbox}[2][]{
    enhanced,
    promptbreak,
    listing only,
    width=\linewidth,
    colback=promptbg,
    colframe=promptorange,
    colbacktitle=promptorange,
    coltitle=white,
    title={#2},
    fonttitle=\bfseries\small,
    boxrule=1pt,
    arc=1mm,
    outer arc=1mm,
    left=2mm,
    right=2mm,
    top=1.5mm,
    bottom=1.5mm,
    listing options={
        basicstyle=\ttfamily\scriptsize,
        breaklines=true,
        breakatwhitespace=false,
        columns=fullflexible,
        keepspaces=true,
        showstringspaces=false,
        tabsize=2,
    },
    #1
}

\title{\textsc{HarnessCompass}: Guiding Automatic Harness Evolution toward Generalizable and Effective Agent Harnesses}
\author{
    Luan Zhang\textsuperscript{\rm 1}\equalcontrib,
    Ruochen Zhou\textsuperscript{\rm 2}\equalcontrib,
    Dandan Song\textsuperscript{\rm 1}\thanks{Corresponding author.},
    Zhengyu Chen\textsuperscript{\rm 3},
    Yuhang Tian\textsuperscript{\rm 1},
    Jun Yang\textsuperscript{\rm 1},
    Huipeng Ma\textsuperscript{\rm 1},
    Chenhao Li\textsuperscript{\rm 1},
    Guangyuan Feng\textsuperscript{\rm 1},
    Xudong Li\textsuperscript{\rm 1},
    Yizhou Jin\textsuperscript{\rm 1},
    Yan Xu\textsuperscript{\rm 1}
}
\affiliations{
    \textsuperscript{\rm 1}Beijing Institute of Technology, China\\
    \textsuperscript{\rm 2}City University of Hong Kong, China \\
    \textsuperscript{\rm 3}Independent, China \\

    \{luan\_zhang, sdd\}@bit.edu.cn\\
    ruochzhou2-c@my.cityu.edu.hk
}

\usepackage{pifont}
\usepackage{pgfplots}
\usepackage{tcolorbox}
\tcbuselibrary{skins}

\begin{document}

\maketitle

\begin{abstract}
Harness design plays a critical role in agent performance by shaping how large language models (LLMs) perceive, reason over, and act within executable environments. Recent work has proposed automatic harness evolution, which iteratively improves the harness from agent--environment interactions. However, existing methods often overfit to the evolution tasks, rely exclusively on trajectory-derived signals, and optimize harness components jointly, causing interference across components. We propose \textbf{\textsc{HarnessCompass}}, a novel automatic harness evolution framework built around \textbf{constrained evolution}, \textbf{proactive feedback}, and \textbf{component-wise optimization}. \textsc{HarnessCompass} first enforces global constraints on evolution, restricting modifications to task-agnostic harness changes that generalize beyond the evolution tasks. It then augments trajectory-derived evidence with proactive first-person feedback from the agent about harness usage, yielding richer signals for evolution. Finally, it decouples the optimization of different harness components before consolidating them into a unified harness, reducing cross-component interference while preserving component synergy. On SWE-bench Verified with GPT-5.4, \textsc{HarnessCompass} improves Pass@1 from 54\% to 66\% in only 5 evolution iterations, outperforming AHE in both effectiveness and evolution efficiency. In addition, the evolved harness transfers effectively to held-out tasks and other models, demonstrating substantially stronger generalization than prior automatic harness evolution methods.  
\end{abstract}


\section{Introduction}
\label{sec:intro}

Large language model (LLM) agents are increasingly deployed in executable environments to fix code repositories, run multi-step terminal workflows, and coordinate tools over long horizons~\citep{swebench,terminalbench,gaia}. 
Beyond the base model, their performance hinges on the \emph{harness}, the software layer that orchestrates how the model interacts with its environment, including the prompts and tools, as well as the middleware, memory, and verification routines that shape what it observes and does~\citep{ahe}. 
Recent work shows that harness design alone can substantially shift agent performance even when the base model is fixed, making it an important lever for improving agents~\citep{trivedy,harnesseng}. 
Because the best harness is model-specific and must be re-tuned whenever the base model changes, this work still falls to developers, who read trajectories, diagnose failures, and revise the harness by hand. As base models advance quickly, this manual loop falls behind, and a widening gap opens between what a model can do and what its harness lets it realize.

This has motivated the development of automatic harness evolution, where a meta-agent iteratively improves the harness based on agent–environment interactions. 
These methods typically follow a search loop, in which a meta-agent analyzes past trajectories generated by the task agent, proposes harness modifications, evaluates the modified harness on benchmark tasks, and iterates~\citep{metaharness,aevo}. 
Such methods search over the entire harness space and report substantial gains on public benchmarks. 
However, a closer look at how these gains arise reveals three limitations that together cap the reliability and generality of current methods. 
First, \ding{182} existing methods overfit to the search tasks. Since the meta agent may revise the harness freely against verifier feedback from a fixed task set, it tends to bake in task-specific shortcuts that lift scores on the search set but fail to transfer. Recent evaluation studies confirm this, reporting that evolved harnesses generalize poorly to held-out tasks once the search and evaluation sets are disjoint, which points to adaptation to specific tasks rather than transferable harness design~\citep{rethinking}.
Second, \ding{183} existing methods rely solely on trajectory-derived signals. The meta agent sees each interaction only from the outside, so it learns that a task failed and where, but not why the agent found the harness hard to use. As a result, it is prone to misattribution, in which genuine harness friction is read as an agent error, or an agent's own reasoning failure is wrongly blamed on a missing tool.
Third, \ding{184} existing methods optimize all harness components jointly. When prompts, tools, middleware, and memory are revised together in one pass, their edits interfere, so the gains each edit brings in isolation tend to cap rather than compound once combined. 

\begin{figure*}[!t]  
  \centering
  \includegraphics[width=0.95\linewidth]{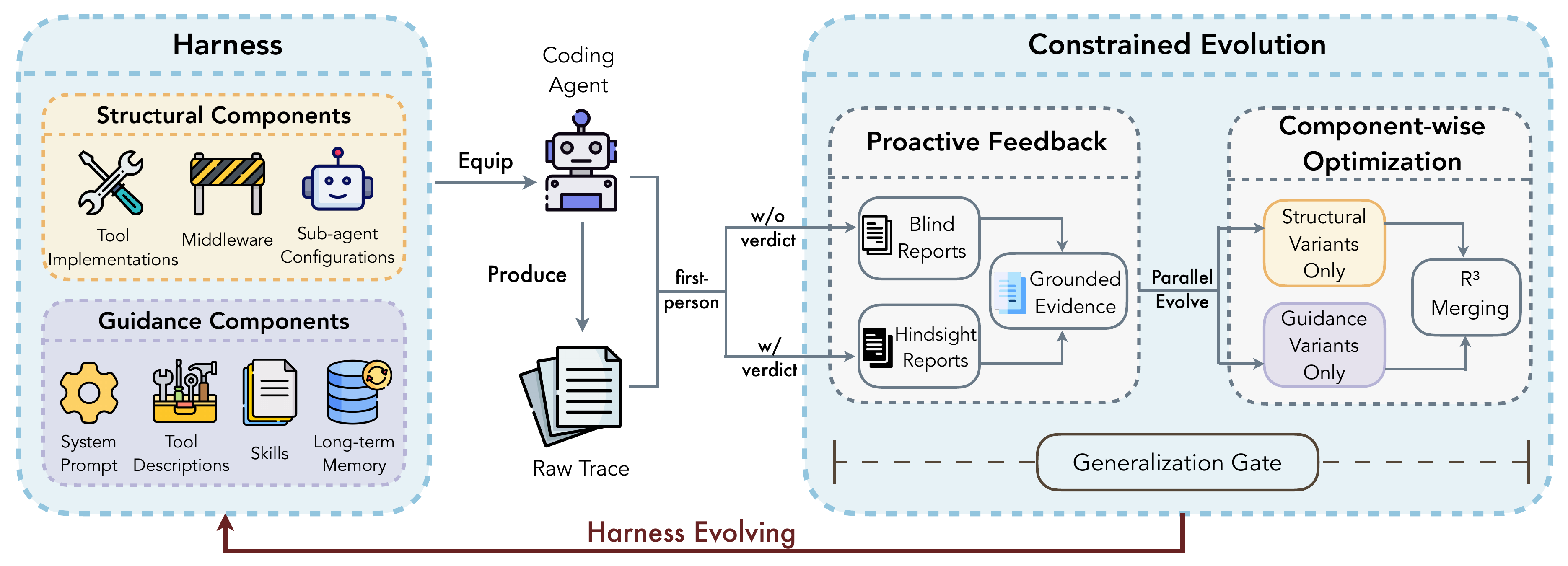}
  \caption{Overview of \textsc{HarnessCompass}. 
  }
  \vspace{-8pt}
  \label{fig:main}
\end{figure*}

These limitations arise at different points in the loop, but they trace back to a single cause. Current methods give the meta agent broad freedom to change the harness while feeding it only external outcomes and letting it revise everything at once. Unrestricted edits invite overfitting, outcome-only evidence invites misattribution, and simultaneous edits invite interference. 
The core challenge, therefore, is not harness evolution itself, but the lack of structure and discipline governing the evolution process.
An effective harness should instead emerge from a disciplined evolution process that constrains what may change, provides the meta-agent with richer evidence for reasoning, and isolates individual modifications before consolidation. We adopt this discipline as the central design goal of our framework. 

Motivated by this goal, we propose \textsc{HarnessCompass} (an overview is provided in Figure~\ref{fig:main}), an automatic harness evolution framework built around three principles that address the three limitations in turn. 
\ding{182} Constrained evolution holds every edit to global constraints. It allows only task-agnostic changes, ruling out hardcoding, keyword matching, and instance-specific rules, so that the framework learns reusable design principles rather than shortcuts fitted to the seen set. 
\ding{183} Proactive feedback adds a signal that traces alone cannot provide, asking the agent to report first-person on its own experience of using the harness, which exposes usage friction and missing capabilities. To keep this reliable, the feedback counts only as candidate evidence. It is grounded against the trajectories before it can drive an edit, so the added signal stays both richer and verifiable.  
\ding{184} Component-wise optimization tunes each harness component on its own track rather than editing everything at once, then merges the validated modifications into a unified harness through a principled integration step. This separation curbs cross-component interference while retaining the synergy between complementary edits.

We evaluate \textsc{HarnessCompass} on SWE-bench Verified. With GPT-5.4 as the code agent, \textsc{HarnessCompass} achieves a substantial improvement in Pass@1, increasing it from 54\% to 66\% after only five evolution iterations. It surpasses AHE, the state-of-the-art automatic harness evolution approach, in both final Pass@1 and evolutionary efficiency. 
More telling is that the evolved harness achieves genuine generalization rather than task-specific optimization, since it carries over to held-out tasks and to other models. This transfer marks a clear gain in generalization over prior automatic harness evolution methods, and it directly mitigates the overfitting to search tasks that has constrained their effectiveness. In summary, our contributions are threefold: 
\begin{itemize} 
    \item We identify three key limitations of existing automatic harness evolution methods, namely search-task overfitting, insufficient feedback for diagnosing harness issues, and interference among jointly optimized components. 
    \item We introduce \textsc{HarnessCompass}, a disciplined harness evolution framework that combines constrained evolution, proactive feedback, and component-wise optimization to produce effective and transferable harnesses. 
    \item We conduct extensive experiments on SWE-bench Verified, demonstrating that \textsc{HarnessCompass} achieves substantial performance gains while improving generalization across unseen tasks and models. 
\end{itemize}

\section{Related Works}
\paragraph{LLM agents and harness design.} 
Deploying an LLM as an agent in an executable environment requires more than a capable base model; it also requires a well-designed harness that modulates how the model perceives state, interacts with tools, and recovers from failures~\citep{zhang2026prunetir}. 
Foundational paradigms such as reasoning-and-acting~\citep{yao2023react} established the interaction loop for LLM-based agents. Building on this foundation, software engineering agents have shown that carefully designed harnesses and interfaces are critical to performance. SWE-agent demonstrates that a purpose-built agent–computer interface substantially improves issue resolution on SWE-Bench~\citep{yang2024sweagent,jimenez2024swebench}, while platforms such as OpenHands unify prompts, tools, and control logic into complete agent scaffolds~\citep{wang2024openhands}. 
Recent evidence further highlights this trend: modifying the agent scaffold alone can yield performance gains comparable to those achieved by upgrading the underlying model. This observation has shifted the field from manually designed scaffolds toward agents capable of autonomously improving their own implementations. Along this direction, the Darwin Gödel Machine views agents as editable software systems and improves them through iterative self-modification and empirical evaluation~\citep{zhang2025dgm}. Similarly, Live-SWE-agent starts from a minimal Bash-only scaffold and autonomously evolves its own implementation at runtime while solving real-world software engineering tasks~\citep{xia2025liveswe}. 
A recent survey organizes this emerging field around what, when, and how an agent evolves\citep{gao2025selfevolving}. However, existing systems largely craft or evolve harnesses without explicit mechanisms for ensuring effectiveness and generality, motivating methods that automatically adapt harnesses while maintaining their performance and transferability. 

\paragraph{Automatic optimization of prompts, tools, and workflows.} 
A parallel line of work replaces manual tuning with automatic search over parts of the agent stack. 
At the prompt level, early works automate instruction optimization by leveraging LLM-based search. OPRO generates improved instructions from prior scores~\citep{yang2023opro}, while TextGrad optimizes compound systems by propagating natural-language feedback~\citep{yuksekgonul2024textgrad}. Recent methods further reduce supervision requirements, with Self-Supervised Prompt Optimization deriving signals solely from output comparisons without ground-truth references~\citep{xiang2025spo}. A recent survey summarizes this rapidly growing area~\citep{ramnath2025apo}.
Beyond prompts, recent work explores automatic tool acquisition and generation for agents. ToolMaker transforms code repositories into callable tools through a self-correcting loop~\citep{wolflein2025toolmaker}, while ToolLibGen organizes automatically generated tools into a maintainable library~\citep{yue2025toollibgen}. 
At the structural level, ADAS searches agent designs in code~\citep{hu2024adas}, while AFlow formulates workflow construction as Monte Carlo tree search over code graphs~\citep{zhang2024aflow}. More broadly, end-to-end agent optimizers further extend this direction~\citep{eversHood2025evolving}. 
Most closely related, automatic harness evolution jointly optimizes the full harness through agent–environment interactions~\citep{ahe}. However, recent evaluation reveals that such methods often overfit search tasks and generalize poorly to disjoint evaluation sets~\citep{rethinking}, highlighting the reliability and generalization gaps addressed in our study. 

\section{{\textsc{HarnessCompass}}}
\label{sec:method}

\textsc{HarnessCompass} is an automatic harness evolution framework that improves the harness of a coding agent while keeping the base model fixed.  
Following prior work~\citep{ahe}, it adopts an iterative closed loop consisting of three stages: evaluating the current harness on search tasks, analyzing the resulting trajectories, and improving the harness based on the analysis. 
What sets \textsc{HarnessCompass} apart is how it disciplines this loop, following the three principles motivated in Section\ref{sec:intro}. First, constrained evolution restricts what the meta agent may edit, so that changes stay task-agnostic and transferable. Second, proactive feedback enriches the evidence the meta agent reasons over with a first-person signal that raw traces do not carry. Third, component-wise optimization optimizes different component classes on separate tracks and then reconciles the surviving edits into one harness, so that the tracks do not interfere. 
Together, these principles transform an evolution loop that would otherwise overfit to search tasks, rely on limited evidence, and suffer from cross-component interference into one that produces effective and generalized harnesses. Algorithm~\ref{alg:loop} shows how the three principles work together within a single iteration.

\begin{algorithm}[t]
\caption{Evolution procedure of \textsc{HarnessCompass}}
\label{alg:loop}
\begin{algorithmic}[1]
\Require initial harness $\mathcal{H}_0$, base model $M$, meta agent $\mathcal{M}$, search tasks $\mathcal{D}$, rollouts per task $k$, iterations $N$
\Ensure evolved harness $\mathcal{H}^{\star}$
\State $\mathcal{H}^{\star} \gets \mathcal{H}_0$
\For{$t = 1$ \textbf{to} $N$}
    \State $\mathcal{T} \gets \Call{Rollout}{M, \mathcal{H}^{\star}, \mathcal{D}, k}$ 
    \Comment{evaluate current harness}
    \State $E \gets \Call{Distill}{\mathcal{T}}$ 
    \Comment{trajectory-based evidence}
    \State $F \gets \Call{GroundSelfFeedback}{\mathcal{M}, \mathcal{T}}$ 
    \Comment{first-person evidence (\textit{see Sec. \textbf{Proactive Feedback}})} 

    \ForAll{track $c \in \{\mathrm{structure},\, \mathrm{guidance}\}$}
        \Comment{component-wise optimization}
        \State $\mathcal{H}^{c} \gets \Call{GatedEvolve}{\mathcal{M}, \mathcal{H}^{\star}, E, F, c}$
        \Comment{constrained evolution (\textit{see Sec. \textbf{Constrained Evolution})}}
        \State $\mathcal{T}^{c} \gets \Call{Rollout}{M, \mathcal{H}^{c}, \mathcal{D}, k}$
    \EndFor

    \State $w \gets \arg\max_{c}\, \mathrm{Pass@1}(\mathcal{T}^{c})$; 
    \enspace $\ell \gets$ \textit{the other track}
    \State $\mathcal{H}_t \gets \mathrm{R}^{3}(\mathcal{H}^{w}, \mathcal{H}^{\ell})$
    \Comment{integrate complementary edits (\textit{see Sec. \textbf{Component-wise Optimization})}}

    \If{$\mathrm{Pass@1}(\mathcal{T}^{w}) > \mathrm{Pass@1}(\mathcal{T})$}
        \State $\mathcal{H}^{\star} \gets \mathcal{H}_t$
        \Comment{accept improved harness}
    \EndIf
\EndFor
\State \Return $\mathcal{H}^{\star}$
\end{algorithmic}
\end{algorithm}

\subsection{Preliminaries}
\label{sec:prelim}

Let $M$ denote the base model, which remains fixed throughout the evolution process, and let $\mathcal{H}$ denote a harness, represented as a set of editable components. Following~\citep{ahe}, a harness consists of seven orthogonal component types. 
These components are divided into two categories: \textbf{structural components}, including tool implementations, middleware, and sub-agent configurations that execute code and control interactions; and \textbf{guidance components}, including the system prompt, tool descriptions, skills, and memory that provide behavioral guidance. Running $M$ with a harness $\mathcal{H}$ on a task $x \in \mathcal{D}$ produces a trajectory $\tau$, also referred to as a raw trace, which records the complete sequence of messages, tool calls, and environment observations. A verifier assigns an outcome $r(\tau)\in{0,1}$ to each trajectory, where $1$ indicates successful task completion. We perform $k$ rollouts per task and evaluate performance using Pass@1, defined as the average success rate over the evaluated tasks. 

The loop evolves a harness over iterations $t=1,\dots, N$. At each iteration $t$, the meta agent $\mathcal{M}$ receives the structured evidence distilled from the round run under $\mathcal{H}_{t-1}$ and produces the next harness $\mathcal{H}_t$. 
We deliberately initialize the process with a minimal harness $\mathcal{H}_0$, consisting only of a single shell command as its tool, a short system prompt, and no middleware, skills, or sub-agents. 
Starting minimal ensures that every component of a later harness $\mathcal{H}_t$ is one that the loop itself introduced and measured against the rollouts of $M$ under $\mathcal{H}$, rather than one inherited from a strong starting point whose contribution we could not isolate.

\subsection{Constrained Evolution}
\label{sec:constrained}

Constrained evolution counters overfitting by making task-specific edits inadmissible from the outset, so the meta-agent can raise the score only through changes that also transfer to unseen tasks. We implement it as a global \textbf{\emph{generalization gate}} that every candidate edit must pass before being applied. The gate covers all seven component types and imposes two requirements on each edit, one on what the edit may express and one on where it may reside.

The first requirement governs the content of an edit. Here, the gate rejects any edit that mentions a specific task instance, test function, or private symbol under evaluation, as well as any code branch that triggers on tokens appearing exclusively in particular tasks. What it admits is the complement of these, a reusable decision criterion paired with an applicability condition that the agent can evaluate on a task it has never seen. This requirement shifts the target of evolution from memorized fixes toward transferable design principles, which are exactly what overfitted harnesses fail to capture. 

The second requirement governs where an edit may be placed. We separate edits into two categories:
\begin{itemize}
    \item \textbf{Capability edits} add new executable functionality, such as running tests or installing missing packages. These edits must be implemented as executable code within middleware, tools, or sub-agents.
    \item \textbf{Guidance edits} only provide behavioral instructions to the agent. These edits must be placed in the system prompt or memory rather than executable components.
\end{itemize}

This separation prevents guidance from being encoded as executable logic. Unlike capability code, guidance does not define a reliable execution condition: it may be triggered on unrelated tasks or fail to activate when needed. Allowing such edits would therefore reintroduce the task-specific behaviors that the generalization gate is designed to eliminate. 

Together, the two requirements ensure that every edit is both task-agnostic and appropriately placed, guiding harness evolution toward reusable engineering principles rather than task-specific optimizations. 

\subsection{Proactive Feedback}
\label{sec:feedback}

Proactive feedback enriches the evidence the meta agent reasons over. Prior automatic harness evolution methods~\citep{ahe} observe each interaction only from an external perspective, so they show that a task failed and where, but not why the code agent struggled to use the harness. We close this gap in three steps. We first ask the code agent itself to report on its own experience of using the harness, since only the code agent knows where it got stuck. We then check each report against the trajectory and keep only what the trajectory actually supports. Finally, we turn the surviving reports into evidence the meta agent can use, alongside the trajectory evidence it already has.

\paragraph{Eliciting first-person feedback.}
After the verifier scores a round, we collect first-person feedback from each failed task using the same base model $M$ that generated the trajectory. We issue two queries and reconcile their outputs: 

\begin{itemize}
    \item \textbf{Blind report.} We provide the trajectory without revealing the verdict and ask the code agent where the harness hindered its execution and what capabilities it wished the harness had provided. This design avoids hindsight bias: without knowing the outcome, the agent is less likely to rationalize the failure from a known result. 

    \item \textbf{Hindsight report.} We reveal the verdict and ask the code agent to diagnose the failure, assigning it to one of four causes: \textit{the harness}, \textit{the agent's own reasoning}, \textit{task ambiguity}, or \textit{the environment}. This attribution isolates harness-related issues from uncontrollable factors, preventing the evolution process from optimizing for failures that the harness cannot fix. 
\end{itemize}

We then reconcile the blind and hindsight reports into unified feedback items, recording their agreement. The resulting feedback serves as additional evidence for harness evolution alongside the trajectory-derived evidence.

\paragraph{Grounding against trajectories.}
A first-person report is merely a claim, and a code agent may attribute its own failures to the harness. We therefore retain a feedback item only when it is supported by the trajectory. For complaints about existing components, the trajectory must directly reveal the underlying difficulty. For requests for new capabilities, which cannot be observed directly in the trajectory, the trajectory must instead demonstrate a concrete gap that the proposed capability would address. Unsupported feedback is discarded, removing both spurious complaints and failures incorrectly attributed to the harness. The remaining items form the \textbf{\emph{grounded evidence}} consumed by the meta agent during evolution.

\paragraph{Aggregation and use.}
We aggregate the surviving reports into structured evidence that the meta agent can directly use. 
This aggregation follows a deterministic procedure without any additional model calls. We group reports by the harness component they implicate and assign each group a confidence score based on three factors: the level of agreement among reports, the number of distinct tasks affected, and the consistency between blind and hindsight reports. 
These grouped reports are provided to the meta agent as an additional evidence source rather than a replacement for trajectory evidence. 
Any edit prompted by this evidence is held to the same standard as any other edit, so it must still be supported by the trajectory and is reverted in the next round if it does not help. In this way, proactive feedback enriches the evidence available to the meta agent while remaining within the constraints imposed by constrained evolution.

\subsection{Component-wise Optimization}
\label{sec:component}
 
Component-wise optimization counters interference caused by jointly modifying full harness components. We avoid this by optimizing components on separate tracks and reconciling the results only afterward. This process consists of two steps: independently optimizing each track and integrating the resulting modifications into a unified harness.

\paragraph{Running the tracks independently.}
In each round, we evolve two harness variants in parallel, with each variant restricted to one of the two component groups defined by constrained evolution. One variant focuses on structural components, including middleware, tool implementations, and sub-agents that execute code and control interactions. The other focuses on guidance components, including the system prompt, skills, tool descriptions, and memory that provide behavioral guidance. Since the two variants operate on disjoint component spaces, their updates remain isolated within each round. We evaluate both variants and select the one with the higher Pass@1 as the winner.

\paragraph{Merging the tracks -- R\textsuperscript{3}.}
The winner is only half of the round's progress, since the loser edited a different component group and may still contain useful improvements. To preserve these gains without inheriting regressions, we perform a three-step merge starting from the winner as the base. We call this procedure R\textsuperscript{3}, corresponding to its three stages: \emph{Revision}, \emph{Recombination}, and \emph{Refinement}. \textbf{Revision} examines each loser edit and retains only those that are independently beneficial and neither duplicate nor conflict with the winner, discarding regressions, task-specific rules, and uncertain modifications. \textbf{Recombination} applies the retained edits to the winner-based harness, with the winner taking precedence in any file-level conflict. \textbf{Refinement} finally removes redundant edits that do not conflict at the file level but implement overlapping functionality, retaining the modification associated with the more reliable component.

\section{Experiments}
Our experiments answer three following questions in turn.
\begin{tcolorbox}[colback=gray!8, colframe=gray!30, boxrule=0.4pt, arc=1mm,
  left=6pt, right=6pt, top=3pt, bottom=3pt, boxsep=2pt, enhanced,
  drop shadow={gray!35}]
\textbf{(Effectiveness)} Does \textsc{HarnessCompass} evolve a better harness than prior automatic harness evolution, and at what cost in iterations?\\[2pt]
\textbf{(Attribution)} Does each of the three principles contribute, and how do they trade off?\\[2pt]
\textbf{(Generalization)} Does the evolved harness achieve generalization across held-out tasks and base models rather than overfit to the search tasks? 
\end{tcolorbox}

\subsection{Setup}

\paragraph{Benchmark.}
We perform evolution on SWE-Bench Verified~\citep{swebench}, a human-validated subset of 500 SWE-Bench tasks covering real-world issues from Python repositories.  
To explicitly evaluate generalization, we randomly sample 50 tasks from SWE-Bench Verified as the evolution set and expose the harness evolution process only to these tasks. The remaining 450 tasks are kept unseen and used solely for evaluation. We report results on both splits: performance on the evolution set reflects optimization progress, whereas performance on the held-out set reveals whether the evolved harness transfers beyond the search tasks. 

\paragraph{Metrics.}
Following the setup of AHE~\citep{ahe}, we report Pass@1, the mean binary success rate over $k$ rollouts per task, because a useful harness should enable the agent to solve tasks on its first attempt. We aggregate results across all $500$ tasks using a task-level average, where each task contributes equally to the overall \emph{Total} score. Alongside Pass@1, we report the number of evolution \emph{Turns} required by each method to obtain its final harness, since each turn entails evaluation of the evolved harness, and efficient evolution should reach a strong harness in fewer iterations. 

\paragraph{Models.}
The code agent, the trajectory analyzer, the feedback agent, and the meta agent all share a base model, GPT-5.4~\citep{openai2025gpt5}. Sharing one model ensures that any measured gain is attributable to harness evolution rather than a stronger meta agent or analyzer. 
We run all agents in \textbf{non-thinking mode} to isolate the effect of harness design, as extended reasoning provides an additional performance source that could mask gains from harness evolution. 
For the cross-model study, we freeze the harness evolved with GPT-5.4 and re-evaluate it on Claude-Sonnet-4.6~\citep{anthropic2025claudesonnet45}, again in non-thinking mode, without any further evolution. 

\paragraph{Baselines.}
We compare against two baselines under the same experimental settings. The first is the \emph{harness-free seed} $\mathcal{H}_0$, a minimal harness with only a shell command as its tool and no middleware, skills, or sub-agents, which measures the base model's performance before evolution. The second is \emph{automatic harness evolution}, represented by AHE~\citep{ahe}, a state-of-the-art harness evolution method.

\subsection{Main Results}

\begin{table}[t]
\centering

\resizebox{\columnwidth}{!}{%
\begin{tabular}{lcccc}
\toprule
Method & Sample & Turns & Held-Out & Total \\
\midrule
$\mathcal{H}_0$ (seed) & $54.0\%$ & $0$ & $51.6\%$ & $51.8\%$ \\
AHE & $63.0\%$ & $20$ & $54.7\%$ & $55.5\%$ \\
\textsc{HarnessCompass} & $\mathbf{66.0\%}$ & $\mathbf{5}$ & $\mathbf{60.4\%}$ & $\mathbf{61.0\%}$ \\
\bottomrule
\end{tabular}%
}

\caption{Main results on SWE-Bench Verified with GPT-5.4 (non-thinking mode). All columns are Pass@1: \emph{Sample} on the $50$-task evolution set, \emph{Held-Out} on the $450$ disjoint tasks never seen during evolution, and \emph{Total} the task-weighted score over all $500$ tasks. \emph{Turns} is the number of evolution iterations to reach the reported harness. Best in bold.}
\label{tab:main}
\end{table}

\begin{figure*}[t]  
  \centering
  \includegraphics[width=0.95\linewidth]{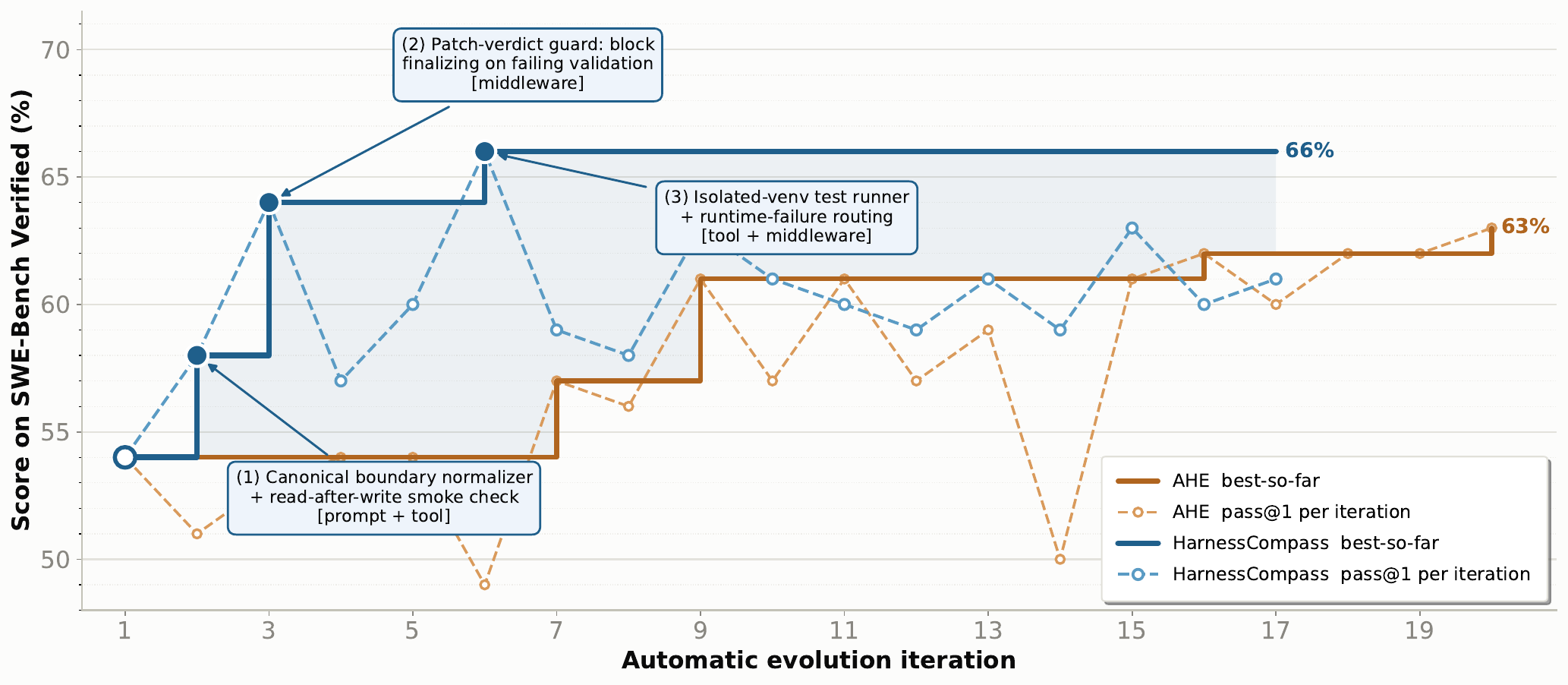}
  \caption{Pass@1 across evolution iterations on the SWE-Bench Verified Sample. 
\textsc{HarnessCompass} reaches its peak performance of $66\%$ at iteration~$5$, whereas AHE approaches a lower plateau only after roughly $20$ iterations.} 
  \vspace{-8pt}
  \label{fig:evo}
\end{figure*}

\paragraph{\textsc{HarnessCompass} evolves a stronger harness in far fewer turns.} Table~\ref{tab:main} presents the main comparison. Starting from the seed harness at $54.0\%$, \textsc{HarnessCompass} improves Pass@1 on the evolution sample to $66.0\%$, outperforming AHE by $3$ points while requiring only $5$ evolution turns compared with AHE's $20$. 
The gap widens on the held-out tasks, where \textsc{HarnessCompass} scores  $60.4\%$ against AHE's $54.7\%$, so its task-weighted total over $500$ tasks rises to $61.0\%$ against AHE's $55.5\%$ and the seed's $51.8\%$. Because all role agents share one base model and both automatic methods start from the identical seed $\mathcal{H}_0$, this margin isolates the effect of the edits the loop commits to the harness. 
The gap further widens on held-out tasks, where \textsc{HarnessCompass} achieves $60.4\%$ Pass@1 versus AHE's $54.7\%$. Its task-level average over all $500$ tasks reaches $61.0\%$, surpassing AHE's $55.5\%$ and the seed harness's $51.8\%$. Since all agents share the same base model and both methods start from the same $\mathcal{H}_0$, the gains reflect improvements from harness evolution.

\paragraph{The advantage is more evident on held-out tasks.}
The performance gap between the two methods is substantially larger on held-out tasks ($5.7$ points) than on the evolution sample ($3.0$ points), highlighting the stronger generalization of \textsc{HarnessCompass}. This improvement stems from the \textit{generalization gate}, which enforces task-agnostic edits and allows improvements discovered on the evolution set to generalize to unseen tasks. Beyond improving generalization, the gate also improves evolution efficiency by preventing the loop from spending iterations on task-specific adaptations that fail to generalize. Overall, constrained evolution enables both stronger generalization and more efficient harness improvement throughout the evolution process.

\begin{table}[t]
\centering

\resizebox{\columnwidth}{!}{%
\begin{tabular}{lcccc}
\toprule
Configuration & Sample & Turns & Held-Out & Total \\
\midrule
$\mathcal{H}_0$ (seed)          & $54.0\%$ & $0$  & $51.6\%$ & $51.8\%$ \\
$+$ Generalization Gate         & $62.0\%$ & $2$  & $58.4\%$ & $58.8\%$ \\
$+$ Proactive Feedback          & $66.0\%$ & $12$ & $55.8\%$ & $56.8\%$ \\
$+$ R\textsuperscript{3} Integration & $\mathbf{66.0\%}$ & $\mathbf{5}$ & $\mathbf{60.4\%}$ & $\mathbf{61.0\%}$ \\
\bottomrule
\end{tabular}%
}
\caption{Ablation on \textsc{HarnessCompass}, adding one principle at a time on top
of the seed. \emph{Sample} is Pass@1 on the $50$-task evolution set, \emph{Held-Out}
on the $450$ disjoint tasks, \emph{Total} the task-weighted Pass@1 over all $500$, and
\emph{Turns} the iterations used. Each principle is cumulative with those above it.}
\label{tab:ablation}
\end{table}

\begin{table}[t]
\centering

\resizebox{\columnwidth}{!}{%
\begin{tabular}{lcccc}
\toprule
Base model & Harness & Sample & Held-Out & Total \\
\midrule
\multirow{2}{*}{GPT-5.4}
 & $\mathcal{H}_0$              & $54.0\%$ & $51.6\%$ & $51.8\%$ \\
 & Ours & $\mathbf{66.0\%}$ & $\mathbf{60.4\%}$ & $\mathbf{61.0\%}$ \\
\midrule
\multirow{2}{*}{Claude-Sonnet-4.6}
 & $\mathcal{H}_0$              & $68.0\%$ & $70.2\%$ & $70.0\%$ \\
 & Ours & $\mathbf{76.0\%}$ & $\mathbf{73.6\%}$ & $\mathbf{73.8\%}$ \\
\bottomrule
\end{tabular}%
}
\caption{Cross-model transfer results. Here, \emph{Ours} refers to \textsc{HarnessCompass}. The harness evolved with GPT-5.4 is frozen and evaluated on Claude-Sonnet-4.6 in non-thinking mode without further evolution. Each base model is compared against its bash-only seed harness. All columns report Pass@1: \emph{Sample} on the $50$-task evolution set, \emph{Held-Out} on the $450$ disjoint tasks, and \emph{Total} over all $500$ tasks.}

\label{tab:crossmodel}
\end{table}

\definecolor{gptseed}{HTML}{A8C5DA}
\definecolor{gptours}{HTML}{1F5C8B}
\definecolor{cldseed}{HTML}{E9C46A}
\definecolor{cldours}{HTML}{BC6C25}

\begin{figure*}[t]
\centering
\begin{tikzpicture}
\begin{axis}[
  width=\textwidth, height=5cm,
  ybar=0.5pt,
  bar width=6.5pt,
  ymin=0, ymax=105,
  ylabel={Pass@1 (\%)},
  ylabel style={font=\small, yshift=-8pt},
  ytick={0,20,40,60,80,100},
  symbolic x coords={django,sympy,sphinx,mpl,sklearn,astropy,pydata,pytest,pylint,psf,mwaskom,pallets,Total},
  xtick=data,
  x tick label style={font=\scriptsize},
  y tick label style={font=\scriptsize},
  enlarge x limits=0.05,
  tick align=inside,
  ymajorgrids=true,
  grid style={gray!25},
  area legend,
  legend style={
    at={(0.5,1.03)}, anchor=south,
    legend columns=4, font=\scriptsize,
    draw=none, /tikz/every even column/.append style={column sep=8pt},
  },
]
\addplot[fill=gptseed, draw=gptseed!70!black] coordinates {
 (django,59.3) (sympy,47.3) (sphinx,35.2) (mpl,38.2) (sklearn,73.4) (astropy,36.4)
 (pydata,59.1) (pytest,52.6) (pylint,15.0) (psf,12.5) (mwaskom,0.0) (pallets,100)
 (Total,51.8)};
\addplot[fill=gptours, draw=gptours!70!black] coordinates {
 (django,70.1) (sympy,52.0) (sphinx,58.0) (mpl,44.1) (sklearn,68.8) (astropy,40.9)
 (pydata,63.6) (pytest,68.4) (pylint,30.0) (psf,12.5) (mwaskom,25.0) (pallets,100)
 (Total,61.0)};
\addplot[fill=cldseed, draw=cldseed!70!black] coordinates {
 (django,73.2) (sympy,68.0) (sphinx,65.9) (mpl,61.8) (sklearn,93.8) (astropy,45.5)
 (pydata,81.8) (pytest,84.2) (pylint,30.0) (psf,12.5) (mwaskom,50.0) (pallets,100)
 (Total,70.0)};
\addplot[fill=cldours, draw=cldours!70!black] coordinates {
 (django,78.8) (sympy,73.3) (sphinx,61.4) (mpl,64.7) (sklearn,93.8) (astropy,63.6)
 (pydata,86.4) (pytest,73.7) (pylint,30.0) (psf,12.5) (mwaskom,50.0) (pallets,100)
 (Total,73.8)};
\legend{GPT-5.4 $\mathcal{H}_0$, GPT-5.4 Ours, Claude-Sonnet-4.6 $\mathcal{H}_0$, Claude-Sonnet-4.6 Ours}
\end{axis}
\end{tikzpicture}
\caption{Per-repository Pass@1 on the full $500$-task SWE-Bench Verified benchmark for GPT-5.4 and Claude-Sonnet-4.6. Each model compares the bash-only seed against its best evolved \textsc{HarnessCompass} harness. Repositories are ordered by task count, from django ($231$) to pallets ($1$); the rightmost group shows the task-level average Pass@1 across all $500$ tasks.} 
\label{tab:perrepo}
\end{figure*}
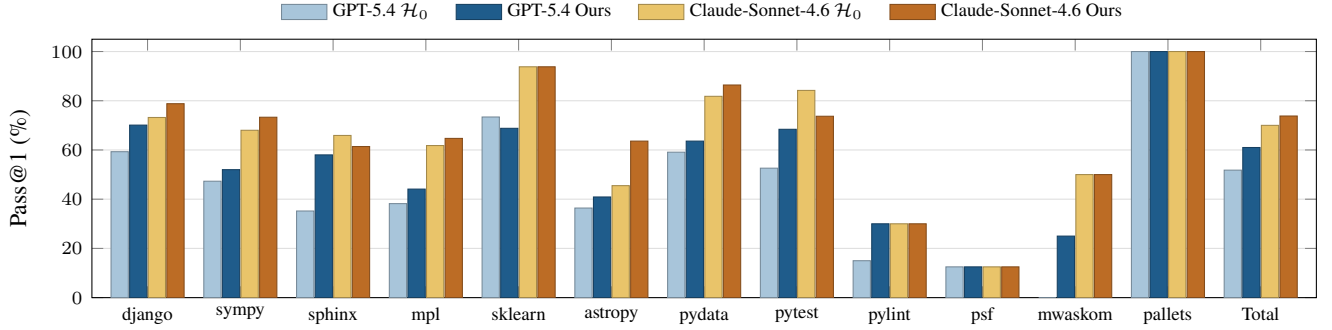

\subsection{Ablation Study}
\label{sec:exp-ablation}
 
To attribute the gains to the three principles rather than the evolution loop as a whole, we conduct a progressive ablation starting from the seed harness and introducing one principle at a time while holding the loop fixed. We first add the generalization gate, followed by proactive feedback and finally R\textsuperscript{3} integration, so that each row measures the marginal contribution of one principle on top of the previous ones.

\paragraph{Each principle contributes, and their effects are complementary.} 
Table~\ref{tab:ablation} shows that the generalization gate alone improves the seed from $54.0\%$ to $62.0\%$ in only two turns, while increasing held-out Pass@1 from $51.6\%$ to $58.4\%$. This confirms that constraining edits to task-agnostic changes is crucial for transferring improvements beyond the evolution set. 
Adding proactive feedback further raises Pass@1 on the evolution set to $66.0\%$, as the richer first-person signal reveals improvement opportunities that trajectory outcomes alone cannot expose. 
However, without safeguards, this gain comes at the cost of generalization, with held-out Pass@1 dropping to $55.8\%$ and turns increasing to $12$, suggesting that feedback-driven edits may overfit to the evolution set.  R\textsuperscript{3} integration resolves this tension by maintaining the $66.0\%$ sample score while restoring held-out Pass@1 to $60.4\%$ and reducing turns to $5$. By reconciling the two component tracks, R\textsuperscript{3} preserves complementary gains while removing redundant or non-transferable edits. 
Together, the three principles play complementary roles: the gate promotes transfer, feedback expands the optimization signal, and R\textsuperscript{3} preserves generalization while improving evolution efficiency.

\subsection{Generalization to Held-Out Tasks and Base Models}
\label{sec:exp-eff}

The held-out column of Table~\ref{tab:main} establishes the first dimension of generalization by showing that, under the same base model, the evolved harness transfers to unseen tasks and substantially narrows the sample-to-held-out gap. 
A harder question is whether the harness also transfers across base models, as a harness optimized for one model's behaviors may not benefit another. To evaluate this, we freeze the harness evolved with GPT-5.4 and re-evaluate it on Claude-Sonnet-4.6 in non-thinking mode without further evolution, comparing each base model against its bash-only seed.

\paragraph{The frozen harness improves a model it was never evolved on.}
Table~\ref{tab:crossmodel} shows that the evolved harness transfers across base models. On Claude-Sonnet-4.6, the frozen harness improves total Pass@1 from $70.0\%$ to $73.8\%$, with gains on both the evolution set (from $68.0\%$ to $76.0\%$) and the held-out split (from $70.2\%$ to $73.6\%$), despite the harness being evolved solely from GPT-5.4. This result suggests that \textsc{HarnessCompass} learns reusable engineering improvements rather than model-specific or task-specific fixes.

\subsection{Evolution Efficiency}

We look beyond the final performance to examine how the two loops evolve, since the turn counts in Table~\ref{tab:main} summarize a trajectory that is itself informative. Figure~\ref{fig:evo} traces per-iteration Pass@1 for \textsc{HarnessCompass} and AHE. 

\paragraph{\textsc{HarnessCompass} climbs faster and settles higher.}
The curve rises steeply and reaches its peak at around iteration~$6$, whereas AHE improves slowly and levels off below it after about $20$ iterations. 
Two mechanisms explain this behavior. The generalization gate keeps the edits transferable, so the harness climbs without the give-back that per-task fixes cause when they revert, and R\textsuperscript{3} folds the losing track's useful edits back into each round, so each iteration compounds rather than discarding half of its exploration. As a result, \textsc{HarnessCompass} not only achieves higher final performance but gets there in roughly one quarter of AHE's iterations.

\subsection{Where the Gains Land Across Repositories}

Consider that the aggregate Total, conceals a heterogeneous benchmark, since SWE-Bench Verified spans twelve repositories whose task counts range from $231$ down to a single task. To see whether the evolved harness helps broadly or merely lifts a few large repositories, we break Pass@1 down per repository for both base models, comparing each seed against its evolved harness in Figure~\ref{tab:perrepo}.

\paragraph{The gains are broadly distributed and concentrated on harness-sensitive repositories.}
On GPT-5.4 the evolved harness improves or matches every repository with more than a handful of tasks, and the largest gains fall exactly where the seed harness is weakest and the task structure is most harness-sensitive. Sphinx-doc jumps from $35.2\%$ to $58.0\%$, pytest-dev from $52.6\%$ to $68.4\%$, and django which is the largest repository from $59.3\%$ to $70.1\%$. Because these repositories exercise the multi-step edit-and-verify loops that the evolved tools and middleware target, the harness has the most room to help there, whereas on scikit-learn, where the seed harness is already strong at $73.4\%$, the score barely moves. Moreover, the same pattern holds on Claude-Sonnet-4.6, where the frozen harness lifts django, sympy, matplotlib, astropy, and pydata, with astropy rising from $45.5\%$ to $63.6\%$, which confirms that the repository-level gains transfer across base models rather than reflecting model-specific behaviors. 

\section{Conclusion}
Observing that existing automatic harness evolution can overfit to search tasks, rely on trajectory-level signals, and introduce interference when jointly optimizing harness components, we propose \textsc{HarnessCompass}, which guides harness evolution through constrained evolution, proactive feedback, and component-wise optimization. 
These three principles ensure that harness modifications remain task-agnostic, enrich trajectory evidence with first-person feedback, and isolate component tracks to reduce cross-component interference. As a result, \textsc{HarnessCompass} achieves stronger effectiveness and generalization than prior methods. Our findings suggest that disciplined evolution is as important as search itself, motivating future work toward harnesses that are both effective and broadly generalizable.

\appendix
\makeatletter
\newif\ifpromptstandalone
\ifx\@nodocument\relax\promptstandalonefalse\else\promptstandalonetrue\fi
\makeatother

\ifpromptstandalone
  \documentclass[10pt,letterpaper,twocolumn]{article}
  \usepackage{times}          
  \usepackage{booktabs}       
  \usepackage{graphicx}       
  
  \makeatletter
  \newcommand{\promptstub}[1]{\@bsphack
    \protected@write\@auxout{}{\string\newlabel{#1}{{0}{1}}}\@esphack}
  \makeatother
  \providecommand{\citep}[1]{[#1]}
  \begin{document}
  \promptstub{sec:method}\promptstub{sec:prelim}\promptstub{sec:constrained}
  \promptstub{sec:feedback}\promptstub{sec:component}\promptstub{alg:loop}
  \promptstub{tab:main}\promptstub{tab:perrepo}
  \appendix
\fi

\section*{Appendix}
\section{Prompts}
\label{app:prompts}

Appendix A presents the prompts implementing the harness components introduced in \textsc{HarnessCompass}. 
We organize them according to the design principle they realize: the seed harness $\mathcal{H}_0$, the generalization gate for constrained evolution, the four-stage first-person reflection pipeline for proactive feedback, and the track constraints together with the R\textsuperscript{3} integrator for component-wise optimization. 
We conclude with the trajectory-analysis prompts inherited from the standard evaluate--analyze--improve loop. 

All prompts are reproduced verbatim. Jinja placeholders (\texttt{\{\{ ws \}\}}, \texttt{\{\{ iteration \}\}}) and Python format fields (\texttt{\{pre\_json\}}, \texttt{\{loser\_diff\}}) are instantiated by the harness at runtime, while non-ASCII characters have been transliterated for typesetting.

\subsection{Seed Harness $\mathcal{H}_0$}
\label{app:seed}

The seed harness is deliberately minimal, consisting only of a shell tool and the system prompt shown below, with no middleware, skills, sub-agents, or long-term memory. Consequently, every component in a later harness $\mathcal{H}_t$ is introduced by the evolution loop itself. 

\begin{strip}
\begin{promptbox}{Seed system prompt (code agent, $\mathcal{H}_0$)}
You solve software tasks in a non-interactive setting. Your only tool is
**`run_shell_command`**: use the shell to inspect the repo, edit files, run
builds/tests, and finish the work. Do not ask the user questions.

- Prefer short replies; use the tool for actions.
- Before commands that delete or overwrite important data, state briefly what they do.
- Long-running processes: use `is_background: true` on `run_shell_command` (do not
  use `&` in the command string).

Date: {{ date }}
Username: {{ username }}
Working Dir: {{ working_directory }}
\end{promptbox}
\end{strip}

\subsection{Generalization Gate}
\label{app:gate}

The gate is specified once in the meta agent's system prompt and enforced uniformly across all candidate edits on the seven component surfaces. The first prompt specifies the content constraint (what an edit may express), whereas the second specifies the placement constraint (where an edit may reside), preventing guidance from being embedded in executable code. 

\begin{strip}
\begin{promptbox}{Generalization gate -- content requirement (meta-agent system prompt)}
## 3. Generalization Gate (applies to EVERY component you author)

Everything the evolve agent writes is GLOBAL context applied to tasks you have never
seen. It must encode a *transferable decision criterion*, never the answer to a
specific observed task.

This gate covers ALL of these surfaces, with no exception:
- `systemprompt.md`
- `LongTermMEMORY.md`
- `skills/**` (SKILL.md and scripts)
- `tool_descriptions/*.tool.yaml`
- `tools/**` and `middleware/**` Python
- `sub_agents/**` -- including each sub-agent's `prompt.md` and `agent.yaml` (a
  sub-agent prompt is a second system prompt -- equally capable of carrying task
  answers)
- `code_agent.yaml`
- Any guidance that instructs the code agent what to record into
  `ShortTermMEMORY.md` at runtime (you cannot edit ShortTermMEMORY directly, but you
  must NOT route task-specific seeds into it indirectly via prompt/skill rules)

**HARD BAN -- none of these may appear in ANY surface above:**
- A specific task / instance id (e.g. a `<library>-<number>` benchmark id).
- A specific test function or test-file name (e.g. `test_*`, `tests/.../*.py`).
- A private symbol / function / class / attribute / file path of a *task-under-test*
  codebase.
- Dataset/task/library-specific keyword matching in code -- e.g.
  `if "<repo-or-test-token>" in text:`. Code branches must NOT switch on tokens that
  only appear in specific tasks.
- "Iteration N showed task X needs change Y" recitations -- these are answers, not
  criteria.

**ONLY allowed:** cross-task reusable principles. Each principle MUST carry an
*applicability criterion* the agent can evaluate on an unseen task -- NOT a
conclusion tied to one task.

Litmus test before writing anything, to any surface above: *"Would this still help a
task from a library I have never seen?"* If it names a task/test/symbol/path,
rewrite it as a criterion or DISCARD it. If it cannot be made task-agnostic, it is
not knowledge -- drop it.

WRONG (answer): "widen the `P` format branch to also handle `Q`".
RIGHT (criterion): "When one shared producer feeds a wrong value to several
call-sites, fix the producer; when the value is wrong only in the current local
scenario, keep the fix local and do NOT promote it into a reused primitive. Decide by
counting how many consumers depend on the value and whether the broken invariant is
global or local."
\end{promptbox}
\end{strip}

\begin{strip}
\begin{promptbox}{Generalization gate -- placement requirement (capability vs. advice)}
### Structural Changes Must Be Deterministic, Not Advisory

Middleware, tools, and sub-agents are **structural** surfaces. A structural change
must add capability: perform a concrete action -- triggered by a real execution event
-- and return a result the agent cannot obtain directly. Examples: run the narrowest
reproducer/test and surface its exit code and traceback; detect a missing dependency
from stderr and install it; verify a patch actually applied and its imports resolve;
expose a tool that locates the owning module/symbol.

A structural change must NOT merely inject advisory text ("remember to verify the
symmetric case", "prefer the canonical owner") into the model context. Advice is not
anchored to an observable failure class, so a structural surface that only injects
advice must guess applicability from broad activity signals such as recent shell
commands or edit types. Those signals are shared by many unrelated tasks, making the
component effectively global and allowing it to leak across cases that the
Generalization Gate is meant to keep isolated. If a change's only effect is advice,
it is a **guidance** change, not a structural one: write the criterion to
`systemprompt.md` / `LongTermMEMORY.md`, where it is read once and governed directly
by the gate.

### Writing to Memory

Memory is always-on global context for ALL tasks -- the highest-conflict surface.
Apply the Generalization Gate (Core Principle 3) strictly:
- Each entry = one transferable principle + its applicability criterion. No task ids,
  no task-under-test symbols, no file/test names.
- Before adding, scan existing memory. If the new lesson duplicates or CONTRADICTS an
  existing entry, do NOT stack them -- merge into a single criterion that states
  *when* each side applies. Contradictory per-task rules stacked verbatim (e.g. "fix
  the shared layer" vs "keep the fix local") are a primary cause of conflict; replace
  both with the deciding criterion.
- Prune every iteration: delete task-specific recitations and entries that no longer
  generalize. Fewer, sharper criteria beat many memorized answers.
- Do NOT work around this gate by telling the code agent (via prompt or skill) to
  record task-specific fixes into ShortTermMEMORY at runtime. The gate follows the
  *content*, not the file.
\end{promptbox}
\end{strip}

\subsection{Proactive Feedback}
\label{app:feedback}

Proactive feedback is used as a four-stage pipeline executed for each failed task. The first three calls are performed by the same base model $M$ that generated the trajectory, ensuring that reflection remains genuinely first-person. The fourth call is handled by the analyzer model, which serves as an external checker. All four calls are grounded in the same trajectory, while only the blind-pass reflection is withheld from the execution verdict.

\paragraph{Step 1: blind report.} The verdict is withheld, preventing the reflection from being retrospectively rationalized based on a known outcome. 

\begin{strip}
\begin{promptbox}{Blind (pre-verdict) first-person reflection}
You are the SAME coding agent whose trajectory is shown above. You have just finished
the task and do NOT yet know whether you passed or failed.
Reflect ONLY on how usable the HARNESS was for you (tools, middleware, skills,
long-term memory, system prompt). Report BOTH:
  (a) friction with EXISTING harness features (kind=improve_existing), and
  (b) capabilities you WISHED EXISTED but did not -- a tool/middleware/skill/sub-agent
      the harness does not currently provide that would have helped you
      (kind=new_capability). For (b), describe the missing capability concretely even
      though it is not in the trace yet.
Do not solve the task; assess the harness.

Output ONE JSON object and nothing else:
{
  "phase": "pre",
  "outcome_visible_to_agent": false,
  "harmful_or_missing_harness_features": [
    {"kind": "improve_existing|new_capability",
     "component": "<middleware|tool_impl|tool_desc|skill|prompt|memory|sub_agent>",
     "friction": "...", "desired_change": "...",
     "trace_refs": ["turn_x"], "severity": 1}
  ],
  "one_change_that_would_have_improved_pass_at_1": "...",
  "risks_of_that_change": "..."
}
\end{promptbox}
\end{strip}

\paragraph{Step 2: hindsight report.} The verifier verdict and its output are appended to this prompt. 

The \texttt{failure\_attribution} field enables us to discard the feedback for a task when the agent itself acknowledges that the failure was not caused by the harness. 

\begin{strip}
\begin{promptbox}{Hindsight (post-verdict) reflection and failure attribution}
You are the SAME coding agent whose trajectory is shown above. The EXTERNAL verifier
has now run; its verdict and output appear below (you never saw it during the task).
With this hindsight, re-assess the HARNESS usability friction and attribute the
outcome. Report BOTH friction with EXISTING features (kind=improve_existing) AND
capabilities you wished existed but the harness lacks (kind=new_capability).

Output ONE JSON object and nothing else:
{
  "phase": "post",
  "verifier_result": "pass|fail",
  "failure_attribution": "harness|agent_reasoning|task_ambiguity|environment",
  "harmful_or_missing_harness_features": [
    {"kind": "improve_existing|new_capability",
     "component": "<middleware|tool_impl|tool_desc|skill|prompt|memory|sub_agent>",
     "friction": "...", "desired_change": "...",
     "trace_refs": ["turn_x"], "severity": 1}
  ],
  "recommended_harness_component":
    "<middleware|tool_impl|tool_desc|skill|prompt|memory|sub_agent>"
}
\end{promptbox}
\end{strip}

\paragraph{Step 3: reconciliation.} The two reports are merged into an item set, with the agreement label recorded. This label is later used to compute the confidence score of the aggregated pattern. 

\begin{strip}
\begin{promptbox}{Reconciling the blind and hindsight reports}
You earlier produced TWO harness-usability reflections for this same task: a
PRE-verdict one (before you knew the result) and a POST-verdict one (after). They are
shown below. Reconcile them into your final, honest view of harness friction. Keep
only what the trajectory genuinely supports -- EXCEPT new_capability items, which
describe something not yet in the harness and so need only be plausible and motivated
by the trajectory. Preserve each item's kind (improve_existing|new_capability).

PRE_REFLECTION:
{pre_json}

POST_REFLECTION:
{post_json}

Output ONE JSON object and nothing else:
{
  "phase": "final",
  "self_consistency": "pre_post_agree|partial|conflict",
  "final_friction_items": [
    {"kind": "improve_existing|new_capability",
     "component": "<middleware|tool_impl|tool_desc|skill|prompt|memory|sub_agent>",
     "friction": "...", "desired_change": "...",
     "trace_refs": ["turn_x"], "severity": 1}
  ]
}
\end{promptbox}
\end{strip}

\paragraph{Step 4: grounding against the trajectory.} 
This call applies item-specific validation: existing-component complaints require trace evidence, whereas missing-capability requests only need to address gaps revealed by the trace. It assigns surviving items to coarse-grained tracks and reclassifies structural items whose \texttt{desired\_change} remains a prompt instruction as guidance rather than advice encoded in code.

\begin{strip}
\begin{promptbox}{Grounding surviving items against the trajectory and routing to a track}
Below is a coding agent's self-reported harness friction for the task whose trajectory
is shown above. The agent's complaints are CANDIDATE EVIDENCE ONLY -- do not trust them
blindly. Apply the right test per item kind:
  - kind=improve_existing: REQUIRES concrete supporting evidence in the trajectory.
    DROP it if the trace does not support it.
  - kind=new_capability: describes a capability the harness does not have yet, so it
    cannot appear in the trace. KEEP it only if the trajectory shows a gap that such a
    capability would plausibly have addressed; cite the turns showing that gap.
For each SURVIVING item, copy its kind/friction/desired_change/severity verbatim and
cite the supporting turns. Then assign a coarse `track` -- NOT a specific component:
  - track=structural: the friction can be addressed by a deterministic code mechanism
    (middleware, a tool's implementation, or a sub-agent). Choose this whenever code
    could enforce/observe/return the missing thing. Do NOT pick a specific component --
    that decision belongs to the evolve agent downstream.
  - track=guidance: the fix is INHERENTLY a cross-task principle that cannot be
    enforced by code (pure prose guidance / a recorded lesson).
Classify NEUTRALLY by the friction's true nature -- do NOT bias toward either track.
Pick structural ONLY when a deterministic code mechanism can genuinely
enforce/observe/return the missing thing; pick guidance when the fix is inherently a
cross-task principle. A friction phrased as 'tell me'/'make it clearer' is guidance
UNLESS code could actually enforce it. Each item goes to exactly the one track that
truly fits.
CRITICAL REWRITE RULE for track=structural items: the downstream structural variant
may ONLY edit code (middleware / tool implementations / sub_agents) and is FORBIDDEN
from touching system prompt / memory / docs. So you MUST rewrite `desired_change` as a
concrete CODE ACTION with NO prompt/doc wording. Forbidden phrasings in a structural
desired_change: 'tell/remind the agent', 'clarify in the prompt', 'document',
'instruct', 'note that', 'make it clear'. Required style: name the code mechanism +
observable effect -- e.g. 'middleware: on stderr match X, auto-run Y and inject its
result before next turn', 'shell tool: detect missing dependency and return install
hint in tool output', 'sub-agent: validate patch applied & imports resolve, return
pass/fail'. If a structural friction genuinely CANNOT be expressed as a code action
(only as prose advice), then it is NOT structural -- reclassify it as track=guidance
instead. Never leave a structural item whose desired_change reads like a prompt
instruction.

FRICTION_ITEMS:
{items_json}

Output ONE JSON object and nothing else:
{
  "grounded_items": [
    {"kind": "improve_existing|new_capability", "friction": "...",
     "desired_change": "...", "severity": 1, "trace_refs": ["turn_x"],
     "track": "structural|guidance"}
  ]
}
Include only items that pass the test above. If none survive, return an empty list.
\end{promptbox}
\end{strip}

\paragraph{Aggregation and use.} Grouping and confidence scoring follow a fixed procedure without additional model calls. The resulting block is inserted into the meta agent's evolution query and filtered by the track accessible to the current variant. Its header explicitly enforces the subordinate role of this evidence relative to the trajectory. 

\begin{strip}
\begin{promptbox}{Aggregated first-person evidence, as injected into the evolution query}
## Agent-Experience Evidence (code-agent self-reported friction, trace-verified)

**Status of this evidence**: this is the code agent's SUBJECTIVE self-report of harness
friction. It is a KNOWLEDGE REFERENCE / SUPPLEMENT only. The Agent Debugger analysis
(section 4 above) and the raw trajectory are the more OBJECTIVE, primary sources --
when they conflict with this section, trust them over this. Use this section to enrich
your understanding of WHY the harness felt hard to use, not as a standalone mandate.

Candidate-evidence ONLY: each item was self-reported by the code-agent (same model,
blind pre-verdict + post-verdict reconciled) and then verified against the trajectory;
unsupported claims were dropped. You MUST still confirm via trajectory and the
falsification loop.

**Generalize, never copy**: the friction text may mention specific
tasks/symbols/files. Do NOT encode those into any component (no keyword-matching, no
per-task dispatch, no hardcoded names). Extract only the cross-task mechanism -- the
fix must help tasks you have never seen. See the Generalization Gate.

**You choose the component**: each pattern carries only a coarse `track` (structural
or guidance) -- NOT a specific component. Like the Agent Debugger analysis, this gives
you evidence, not a prescription. Within your variant's allowed track you decide the
best component yourself (structural -> middleware / tool_impl / sub_agent; guidance ->
systemprompt / memory / skill / tool_desc).

**HARD TRACK BOUNDARY (never cross it)**: choose freely AMONG the components of your
OWN track, but you may NEVER edit a component of the other track. [...]

**DETERMINISTIC-NOT-ADVISORY** (structural track): if a candidate_fix reads like advice
('surface X', 'tell the agent Y', 'document Z'), do NOT satisfy it by editing system
prompt / memory -- that is guidance dressed up as a fix. Re-express it as a
deterministic mechanism (middleware that auto-injects the missing context/dependency, a
tool whose code now returns the missing result, a sub-agent). A prompt edit that merely
restates the friction tends to be INEFFECTIVE.

- [P-structural] track=structural, kind=improve_existing+new_capability,
  confidence 0.87, 4 task(s), 6 grounded item(s).
  Candidate fix (you choose the component): {candidate_fix}
  tasks: {affected_tasks}
  feedback_ids: {feedback_ids}

If you act on a pattern, cite its feedback_ids in the manifest's
`agent_feedback_evidence` and set `feedback_confidence` to its confidence.
\end{promptbox}
\end{strip}

\subsection{Component-wise Optimization}
\label{app:component}

\paragraph{Track constraints.} Each round runs two meta-agents in parallel on the same evidence, differing only in the mandatory constraint appended to their evolution queries. Since the two constraints partition the seven component types, the resulting variants edit disjoint files and cannot interfere with each other within a round. 

\begin{strip}
\begin{promptbox}{Structural variant constraint}
You MUST focus on STRUCTURAL changes: middleware, tool implementations, or sub-agents.
Do NOT modify system prompt or skill content this iteration. CRITICAL -- the
Generalization Gate applies to ALL structural surfaces: they must be GENERIC mechanisms
that work on tasks you have never seen. FORBIDDEN: any code branch that keyword-matches
task/test/library-specific strings (e.g. `if "<token>" in text`) and any hardcoded
per-task dispatch. If you cannot describe the mechanism without naming a specific task,
it is task-specific hacking -- do not write it.

DETERMINISTIC-NOT-ADVISORY: a structural change must alter what the agent
DETERMINISTICALLY DOES and return an observable result the agent cannot obtain directly
-- e.g. auto-run the narrowest reproducer/test and surface its exit code & traceback,
auto-detect-and-install missing dependencies, deterministically validate that a patch
applied and imports resolve, expose a new tool that locates the owning module/symbol.
If a structural change's ONLY effect is to inject advisory text into the model context
("remember to verify X", "prefer the canonical owner", a decision criterion in prose),
it is NOT a structural change -- it is guidance dressed up as a structural change;
discard it rather than writing it as code.
\end{promptbox}
\end{strip}

\begin{strip}
\begin{promptbox}{Guidance variant constraint}
You MUST focus on GUIDANCE changes: system prompt rules, skill packages, tool
descriptions, or LongTermMEMORY. Do NOT create or modify middleware Python files this
iteration. CRITICAL: every rule you write is global context for tasks you have never
seen. Write only cross-task principles, each with an explicit applicability criterion
(when it applies / when it does not) -- never a recitation of a specific task's fix,
symbol, file path, or test name. Before adding to memory, merge or prune contradictory
per-task rules into a single deciding criterion. Do NOT instruct the code agent to
record task-specific fixes into ShortTermMEMORY. If a lesson cannot be made
task-agnostic, discard it.
\end{promptbox}
\end{strip}

\begin{strip}
\begin{promptbox}{Track boundary self-check (appended to both variants)}
You are one of multiple parallel evolve agents. Each agent is assigned a different
strategy direction.
**Your assigned constraint**: {strategy_hint}
You MUST follow this constraint. Violations will waste this variant's exploration
budget.

**Hard boundary + self-check**: STRUCTURAL variant edits ONLY middleware / tool code /
sub_agents and NEVER systemprompt.md / *MEMORY.md / tool_descriptions / skills;
GUIDANCE variant edits ONLY systemprompt / memory / skills / tool_descriptions and
NEVER middleware / tool code / sub_agents. Within your own side you may freely choose
the component, but you may NOT cross to the other side. Before writing
change_manifest.json, verify every edited file is on your side; if any is on the other
side, REVERT it (or re-express the intent within your side) -- do not report a
cross-boundary edit.
\end{promptbox}
\end{strip}

\paragraph{R\textsuperscript{3} merge.} After evaluating both variants, the merge is performed in a git worktree branched from the winner, making the winner's edits the immutable base and the loser's diff the only candidate material. The following prompt jointly drives Revision, Recombination, and Refinement in a single pass. 

\begin{strip}
\begin{promptbox}{R$^3$ integrator}
# R3 Integration -- combine two evaluated harness variants into one

You are the R3 Integrator. This iteration ran TWO harness variants in parallel, each
constrained to a different track:
- variant {winner_idx} (WINNER, pass@1 {winner_rate}) -- the higher-scoring variant
- variant {loser_idx} (loser, pass@1 {loser_rate}) -- the lower-scoring variant

The two variants edited DIFFERENT, non-overlapping component tracks (one structural:
middleware/tool_impl/sub_agent; the other guidance: prompt/skill/tool_desc/memory).
Winner-take-all throws the loser's work away every round. Your job is to salvage the
loser's genuinely useful, independent changes ON TOP OF the winner -- so the result is
>= winner, never worse.

Your workspace `{ws}/` ALREADY CONTAINS THE WINNER's full changes as the base. Do NOT
undo them. You only ADD selected loser changes and then simplify.

## The loser's changes (candidate material to fold in)
{loser_manifest}

## The loser's diff (what it actually changed)
{loser_diff}

## Per-variant evaluation evidence (which tasks each flipped / regressed)
{eval_evidence}

## Do exactly three steps, in ONE pass, writing into `{ws}/`

### 1. Revision (judge the loser's changes)
Go through the loser's changes ONE BY ONE -- enumerate EVERY change in its
manifest/diff, do not stop after the first obvious one. For each, decide KEEP or DROP
using the evidence, and record your verdict for every change:
- KEEP only if it looks like an INDEPENDENT positive contribution (plausibly helped
  tasks, does not duplicate or contradict a winner change).
- DROP if it looks like a regression source, is task-specific hacking, or you cannot
  tell it helped. When unsure, DROP -- the winner base already stands on its own;
  adding doubtful changes only risks lowering it.
Multiple independent good changes can and should ALL be kept -- do not artificially cap
to one. The loser explored a whole track; salvage everything genuinely useful.

### 2. Recombination (apply kept loser changes onto the winner base)
Apply the KEEP changes into `{ws}/`. Because the tracks are orthogonal they should
rarely touch the same file as the winner. If a loser change DIRECTLY conflicts with a
winner change, the winner wins -- skip the loser one.

### 3. Refinement (Occam's razor -- remove redundancy)
Do NOT wait for redundancy to be obvious -- actively HUNT for it. Concretely: walk EACH
advisory item now in the combined harness (every systemprompt rule, every memory entry,
every tool_description note) and ask: "is this behavior ALREADY enforced
deterministically by some middleware / tool_impl in this same harness?" If yes, that
advisory item is redundant -- DELETE it. Two changes are redundant when they do the
SAME job even in different files with no merge conflict. When redundant, KEEP the more
DETERMINISTIC one and DELETE the advisory one -- regardless of which variant (winner or
loser) it came from.

This keeps prompts/memory from bloating over rounds (a real failure mode in past runs).
Still: only delete on a CONFIRMED duplicate -- if you cannot point to the specific
component that already covers it, keep both. The bar is "I can name what makes this
redundant", not "delete freely".

## Constraints
- Stay within the Generalization Gate: no task/test/symbol-specific content.
- Do NOT modify LLM config, tracer, verifier, or any infrastructure.
- Preserve every capability that is not a confirmed duplicate.
- NOTE ON THE SYSTEM-PROMPT RULE "do not delete ORIGINAL system prompt rules": that
  protects ITERATION-1 baseline rules. In Refinement you MAY delete a rule that THIS
  iteration's variants just added if it is a confirmed duplicate of a more
  deterministic component -- that is the whole point of this step. Never delete
  iteration-1 baseline rules.

## Deliverable
This is an INTEGRATION task, not a normal evolution step. Write the manifest as
`{ws}/integration_manifest.json`:
{
  "base": "winner variant {winner_idx}",
  "kept_from_loser": [ {"change": "...", "why": "..."} ],
  "dropped_from_loser": [ {"change": "...", "why": "..."} ],
  "removed_as_redundant": [ {"change": "...", "kept_instead": "...", "why": "..."} ]
}
Then commit all changes in `{ws}/`.
\end{promptbox}
\end{strip}

\subsection{Change Manifest}
\label{app:manifest}

Every meta agent edit is recorded in a machine-readable manifest, which the next round uses to attribute score changes to individual edits and determine rollbacks. The final two fields link each edit to the first-person evidence that motivated it, enabling us to measure how often proactive feedback leads to committed changes.

\begin{strip}
\begin{promptbox}{Change manifest schema (meta-agent deliverable)}
{
  "iteration": {{ iteration }},
  "changes": [
    {
      "id": "chg-1",
      "type": "new|improvement|rollback",
      "description": "What was changed and why",
      "files": ["relative/to/workspace/file.py"],
      "failure_pattern": "The failure class this addresses",
      "predicted_fixes": ["task-name-a", "task-name-b"],
      "risk_tasks": ["task-name-c"],
      "constraint_level": "middleware|tool_impl|tool_desc|skill|prompt",
      "why_this_component": "Why this component level was chosen over alternatives",
      "agent_feedback_evidence": [],
      "feedback_confidence": 0.0
    }
  ]
}

`agent_feedback_evidence` / `feedback_confidence` (OPTIONAL, only when
Agent-Experience Evidence is present): if this change acts on a self-reported friction
pattern, list that pattern's `feedback_ids` and copy its `confidence`. This is
CANDIDATE EVIDENCE that only SUPPLEMENTS trajectory evidence -- never a substitute. A
change still requires real `trajectory_evidence` and must pass the falsification loop;
leave both empty (`[]` / `0.0`) for changes not driven by agent feedback.
\end{promptbox}
\end{strip}

\section{Per-Repository Results}
\label{app:perrepo}

Figure~\ref{tab:perrepo} visualizes per-repository Pass@1 scores as a bar chart. This section provides the corresponding numerical results. Table~\ref{tab:perrepo-full} reports the Pass@1 of the bash-only seed $\mathcal{H}_0$ and the best evolved \textsc{HarnessCompass} harness for each of the twelve repositories in SWE-Bench Verified, under both base models, along with the corresponding improvements. All results are computed over $k=2$ rollouts on the full $500$-task benchmark. Thus, a repository with $n$ tasks contributes $2n$ trials, and its Pass@1 is quantized in increments of $1/(2n)$; values are rounded to one decimal place. The \emph{Total} row reports the task-weighted mean over all $500$ tasks and therefore matches the totals in Table~\ref{tab:main}. 

\begin{table*}[t]
\centering
\small
\caption{Per-repository Pass@1 (\%) on the full $500$-task SWE-Bench Verified benchmark. $\mathcal{H}_0$ denotes the bash-only seed harness, and \emph{Ours} denotes the best evolved \textsc{HarnessCompass} harness; $\Delta$ reports their difference in percentage points, with positive values highlighted in bold. Repositories are ordered by task count $n$. The harness is evolved once with GPT-5.4, then frozen and evaluated on Claude-Sonnet-4.6 without further evolution; thus, the Claude results measure transfer to a base model unseen during optimization. The overall gain is not uniform: improvements are concentrated in larger repositories, whereas several smaller ones remain unchanged due to their limited task counts.}
\label{tab:perrepo-full}
\resizebox{\textwidth}{!}{%
\begin{tabular}{lrcccccc}
\toprule
& & \multicolumn{3}{c}{GPT-5.4} & \multicolumn{3}{c}{Claude-Sonnet-4.6} \\
\cmidrule(lr){3-5}\cmidrule(lr){6-8}
Repository & $n$ & $\mathcal{H}_0$ & Ours & $\Delta$ & $\mathcal{H}_0$ & Ours & $\Delta$ \\
\midrule
\texttt{django/django} & $231$ & $59.3$ & $70.1$ & $\mathbf{+10.8}$ & $73.2$ & $78.8$ & $\mathbf{+5.6}$ \\
\texttt{sympy/sympy} & $75$ & $47.3$ & $52.0$ & $\mathbf{+4.7}$ & $68.0$ & $73.3$ & $\mathbf{+5.3}$ \\
\texttt{sphinx-doc/sphinx} & $44$ & $35.2$ & $58.0$ & $\mathbf{+22.8}$ & $65.9$ & $61.4$ & $-4.5$ \\
\texttt{matplotlib/matplotlib} & $34$ & $38.2$ & $44.1$ & $\mathbf{+5.9}$ & $61.8$ & $64.7$ & $\mathbf{+2.9}$ \\
\texttt{scikit-learn/scikit-learn} & $32$ & $73.4$ & $68.8$ & $-4.6$ & $93.8$ & $93.8$ & $\pm0.0$ \\
\texttt{pydata/xarray} & $22$ & $59.1$ & $63.6$ & $\mathbf{+4.5}$ & $81.8$ & $86.4$ & $\mathbf{+4.6}$ \\
\texttt{astropy/astropy} & $22$ & $36.4$ & $40.9$ & $\mathbf{+4.5}$ & $45.5$ & $63.6$ & $\mathbf{+18.1}$ \\
\texttt{pytest-dev/pytest} & $19$ & $52.6$ & $68.4$ & $\mathbf{+15.8}$ & $84.2$ & $73.7$ & $-10.5$ \\
\texttt{pylint-dev/pylint} & $10$ & $15.0$ & $30.0$ & $\mathbf{+15.0}$ & $30.0$ & $30.0$ & $\pm0.0$ \\
\texttt{psf/requests} & $8$ & $12.5$ & $12.5$ & $\pm0.0$ & $12.5$ & $12.5$ & $\pm0.0$ \\
\texttt{mwaskom/seaborn} & $2$ & $0.0$ & $25.0$ & $\mathbf{+25.0}$ & $50.0$ & $50.0$ & $\pm0.0$ \\
\texttt{pallets/flask} & $1$ & $100.0$ & $100.0$ & $\pm0.0$ & $100.0$ & $100.0$ & $\pm0.0$ \\
\midrule
\textbf{Total} & $500$ & $51.8$ & $61.0$ & $\mathbf{+9.2}$ & $70.0$ & $73.8$ & $\mathbf{+3.8}$ \\
\bottomrule
\end{tabular}%
}
\end{table*}

\paragraph{The gain is broad but concentrated in large repositories.}
Under GPT-5.4, the evolved harness improves eight of the twelve repositories, remains
unchanged on three, and regresses on one. The improvement is not driven by a handful
of small repositories: \texttt{django}, which alone contributes $231$ of the $500$
tasks, gains $10.8$ points, while \texttt{sympy}, the second largest repository with
$75$ tasks, gains $4.7$ points. Together, these two repositories account for $61\%$ of
the benchmark and therefore contribute substantially to the overall $+9.2$ gain.
This pattern is precisely what constrained evolution is designed to achieve: a
harness that merely memorizes its $50$-task evolution set would not generalize to
substantially improve a $231$-task repository that it rarely encountered during
evolution. 

\paragraph{Small repositories have limited statistical resolution.}
The smallest repositories, \texttt{psf} ($8$ tasks), \texttt{mwaskom} ($2$ tasks),
and \texttt{pallets} ($1$ task), have Pass@1 resolutions of $6.25$, $25$, and $50$
percentage points, respectively. As a result, a single trial flip can produce a
change larger than most underlying effects. The $+25.0$ gain on \texttt{mwaskom}
under GPT-5.4 corresponds to one additional success among four trials, while the
$\pm0.0$ entries only indicate that no trial outcome changed. We include these rows
for completeness but caution against over-interpreting them; the same applies to
\texttt{pallets}, whose single task is solved by both models under every harness. 

\paragraph{Transfer to a different base model is positive but uneven.}
The Claude-Sonnet-4.6 results evaluate the GPT-5.4-evolved harness after freezing. The harness transfers positively in aggregate, yielding a $+3.8$ point gain, and improves \texttt{astropy} by $18.1$ points. Still, the gains are not uniform: \texttt{pytest} drops by $10.5$ points and \texttt{sphinx} by $4.5$ points, despite the seed harness already performing strongly on these repositories for this model ($84.2$ and $65.9$, respectively). This suggests that some mechanisms evolved for GPT-5.4 may be unnecessary for a stronger base model. We therefore consider the aggregate transfer performance the primary result.

\section{Case Study: When Ungated Evolution Rewards Overfitting}
\label{app:case-study}

The aggregate score alone does not reveal whether evolution discovered reusable engineering competence or merely memorized solutions to the tasks seen during evolution. We therefore inspect the artifacts that evolution leaves behind. This provides a qualitative mechanistic check on the role of the Generalization Gate, rather than treating Pass@1 as evidence of generalization by itself.

\subsection{Two kinds of accumulated memory}
\label{app:case-memory}

The contrast between the ungated and gated runs is qualitative yet pronounced. In the ungated run, the accumulated memory is dominated by task-specific recipes. Representative entries include:  

\begin{quote}
\small
\texttt{django\_\_django-13158 regressed because guidance drifted toward
shape-preserving QuerySet.none() rewrites. For combined queries, emptiness must
propagate into combined\_queries ...}

\texttt{sympy\_\_sympy-14711 showed arithmetic protocol bugs must be fixed at
the direct failing operator entry point. If A.x + 0 fails, patch
Vector.\_\_add\_\_ first ...}
\end{quote}

These entries explicitly reference benchmark instances and private implementation details. They can therefore serve as compact answer keys: when the same or a closely related task is encountered, the memory directly indicates where to edit and which behavior to implement. However, their apparent stability does not by itself imply that the harness has learned a robust method; the harness may simply be repeatedly retrieving solutions encoded from its evolution tasks.

In contrast, the gated run records general principles together with explicit applicability conditions. For example:

\begin{quote}
\small
\textbf{Execution-path ownership rule.} Applicability: when several nearby layers
could plausibly be patched, trace the failing value to the function that actually
emits it, and patch that owner rather than an upstream helper or downstream
formatter.

\textbf{Consumer-count scope rule.} Applicability: when choosing between a local
fix and a shared-abstraction change, rewrite the shared helper only when multiple
consumers share the same broken contract; otherwise keep the fix local.
\end{quote}

Neither principle identifies a repository, task number, symbol, or expected patch. Instead, each specifies a decision criterion that can be evaluated on an unseen codebase. The applicability clause is important: it prevents a useful observation from being promoted into an unconditional rule, and makes the boundary of transfer explicit.

\subsection{Interpretation}
\label{app:case-interpretation}

The two artifacts support a straightforward causal interpretation of the score difference. Without the gate, evolution can promote task-specific solutions into always-on global context. This creates a direct path from previously observed tasks to higher scores, but leaves the resulting harness tightly coupled to the evolution set and unlikely to benefit unfamiliar tasks. In effect, ungated evolution trades benchmark-specific memorization for apparent performance gains.

The gate removes this shortcut by prohibiting task identifiers, task-specific symbols, and file paths, while requiring every retained lesson to specify its applicability conditions. The gated artifact therefore encodes method-level knowledge rather than a catalogue of patches. Memory inspection alone does not prove generalization, nor does it imply that every gated rule will transfer successfully. It does, however, rule out the direct alternative explanation for the ungated gain: the observed advantage can be attributed to the nature of the evolved artifact, with the ungated run storing specific recipes and the gated run storing conditional, cross-task decision criteria.

In short, ungated memory answers “how to modify django-13158,” whereas gated memory answers “how to decide which layer to modify when a shared contract is violated.” The former may improve performance on the evolution benchmark while functioning as an answer book; the latter captures the form of knowledge required for transfer to unseen tasks. This case study therefore complements the held-out and cross-model results in Appendix~\ref{app:perrepo}: quantitative evaluation measures whether transfer occurs, while artifact inspection explains why an unconstrained score improvement should not be automatically interpreted as stronger general engineering ability.

\subsection{Future Work}
LLM-based agents still suffer from hallucinations and unreliable intermediate behaviors~\cite{lin2025llm, zhang2025detecting, chakraborty-etal-2025-heal, ma2026actishade}, which can be particularly harmful in long-horizon tasks. Unlike single-step generation, agentic tasks involve extended interactions among reasoning, tool usage, and environment feedback, where an early erroneous action or hallucinated assumption can propagate through subsequent steps and accumulate into a final failure~\cite{yu-etal-2026-infiagent, liang-etal-2026-learning-irrecoverable}. A promising direction is to extend automatic harness evolution toward automatically discovering and mitigating such failure patterns. Specifically, evolved harnesses could learn generalizable rubrics that identify intermediate agent behaviors associated with downstream failures, rather than relying only on outcome signals. Based on these rubrics, the harness could further evolve corrective mechanisms, such as trajectory-level filtering, targeted feedback generation, or reward modification, to prevent error propagation during execution. By reducing accumulated failures and improving the quality of collected trajectories, such harnesses could provide stronger supervision signals for reinforcement learning and enable more reliable training of long-horizon agents.

\ifpromptstandalone\else \fi

\end{document}

\bibliography{aaai2027}

\end{document}